\documentclass[twoside,11pt]{article}

\usepackage{jmlr2e}
\usepackage{hyperref}
\usepackage{listings}
\usepackage[frozencache]{minted}
\usepackage{amsmath}
\usepackage{caption}
\usepackage{booktabs}
\usepackage{amssymb}
\usepackage{pifont}
\usepackage{subcaption}
\usepackage{url}
\usepackage{multirow}
\usepackage{booktabs}
\usepackage{makecell}
\usepackage{enumitem}
\usepackage{array}
\usepackage{cleveref}

\newcommand{\cmark}{\ding{51}}%
\newcommand{\xmark}{\ding{55}}%

\newcommand{\bzero}{\mathbf{0}}
\newcommand{\bx}{\mathbf{x}}

\newcommand{\bz}{\mathbf{z}}
\newcommand{\bw}{\mathbf{w}}

\newcommand{\bp}{\mathbf{p}}

\newcommand{\bh}{\mathbf{h}}
\newcommand{\bW}{\mathbf{W}}
\newcommand{\bI}{\mathbf{I}}

\newcommand\longvar[1]{{\rm{\mathchardef\UrlBreakPenalty=100
\mathchardef\UrlBigBreakPenalty=100\url{#1}}}}

\newtheorem{examp}{Example}

\setminted[python]{fontsize=\small, fontshape=n, xleftmargin=.05\textwidth}

\ShortHeadings{ZhuSuan: A Library for Bayesian Deep Learning}{Shi et al.}
\firstpageno{1}

\begin{document}

\title{ZhuSuan: A Library for Bayesian Deep Learning\thanks{J. Zhu is the corresponding author. S. Sun is now with Department of Computer Science, University of Toronto. }}

\author{\name Jiaxin Shi$^1$ \email shijx15@mails.tsinghua.edu.cn \\
		\name Jianfei Chen$^1$ \email chenjian14@mails.tsinghua.edu.cn \\
        \name Jun Zhu$^1$ \email dcszj@tsinghua.edu.cn \\
       \name Shengyang Sun$^2$ \email ssy@cs.toronto.edu \\
       \name Yucen Luo$^1$ \email luoyc15@mails.tsinghua.edu.cn \\
       \name Yihong Gu$^1$ \email gyh15@mails.tsinghua.edu.cn \\
       \name Yuhao Zhou$^1$ \email zhouyh16@mails.tsinghua.edu.cn \\
       \addr $^1$Department of Computer Science \& Technology, TNList Lab,  CBICR Center \\ 
       \addr $^1$State Key Lab of Intelligent Technology \& Systems \\
       \addr $^2$Department of Electronic Engineering \\
       Tsinghua University \\
       Beijing, 100084, China 
       }

\editor{}

\maketitle

\begin{abstract}
In this paper we introduce ZhuSuan, a python probabilistic programming library for Bayesian deep learning, which conjoins the complimentary advantages of Bayesian methods and deep learning. ZhuSuan is built upon Tensorflow. Unlike existing deep learning libraries, which are mainly designed for deterministic neural networks and supervised tasks, ZhuSuan is featured for its deep root into Bayesian inference, thus supporting various kinds of probabilistic models, including both the traditional hierarchical Bayesian models and recent deep generative models. We use running examples to illustrate the probabilistic programming on ZhuSuan, including Bayesian logistic regression, variational auto-encoders, deep sigmoid belief networks and Bayesian recurrent neural networks. 
\end{abstract}


\section{Introduction}

These years have seen great advances of deep learning~\citep{lecun2015deep} and its success in many applications, such as speech recognition~\citep{hinton2012deep}, computer vision~\citep{krizhevsky2012imagenet}, language processing~\citep{sutskever2014sequence}, and computer games~\citep{silver2016mastering}. One good lesson we learned from the practice is that a deeply architected model can be well fit by leveraging advanced optimization algorithms (e.g., stochastic gradient decent), a large training set (e.g., ImageNet), and powerful computing devices (e.g., GPUs). Although the great expressiveness of deep neural networks (DNNs) has made them a first choice for many complex fitting problems, especially those tasks involving a mapping from an input space to an output space (e.g., classification and regression), the results given by them are usually point estimates. Such a deterministic approach does not account for uncertainty, which is essential in every part of a learning machine (e.g., data, estimation, inference, prediction and decision-making) due to the randomness of the physical world, incomplete information, and measurement noise. It has been demonstrated that deterministic neural networks can be 
vulnerable to adversarial attacks~\citep{szegedy2013intriguing, nguyen2015deep} (partly because of the lack of uncertainty modeling), which may hinder their applications in the scenarios 
where representing uncertainty is of crucial importance, {such as automated driving~\citep{bojarski2016end} and healthcare~\citep{miotto2017deep}.\footnote{Some recent efforts have been made towards improving the robustness of deep neural networks against adversarial samples by doing adversarial training~\citep{szegedy2013intriguing}, reverse cross-entropy training~\citep{RobustDL-RCE2017} or Bayesian averaging~\citep{RobustDL-Bayes2017}.} 
On the other hand, the probabilistic view of machine learning offers mathematically grounded tools for dealing with uncertainty, i.e., Bayesian methods \citep{ghahramani2015probabilistic}. Bayesian methods also provide a theoretically sound approach to incorporating structural bias~\citep{lake2015human} and domain knowledge as prior or posterior constraints~\citep{mei2014robust} to achieve efficient learning with a small number of training samples. Thus it is beneficial to combine the complimentary advantages of deep learning and Bayesian methods, which in fact has been drawing tremendous attention in recent years~\citep{Gal2016Uncertainty}.

Moreover, as most of the success in deep learning comes from supervised tasks that require a large number of labeled data, research in this area is paying more and more attention to unsupervised learning, a long-standing goal of artificial intelligence (AI). Probabilistic models and Bayesian methods are commonly treated as principled approaches to modeling unlabeled data for pure unsupervised learning or its hybrid with supervised learning (aka., semi-supervised learning). 
Recently, the popularity of deep generative models again demonstrates the promise of combining deep neural networks with probabilistic modeling, which has shown superior results in image generation~\citep{kingma2013auto,goodfellow2014generative, radford2015unsupervised}, semi-supervised classification~\citep{kingma2014semi,salimans2016improved} and one-shot learning~\citep{rezende2016one}.

We call such an arising direction that conjoins the advantages of Bayesian methods and deep learning as \textit{Bayesian Deep Learning} (BDL). The scope of BDL covers the traditional Bayesian methods, the deep learning methods where probabilistic inference plays a key role, and their intersection. 
One unique feature of BDL is that the deterministic transformation between random variables can be automatically learned from data under an expressive parametric formulation typically using deep neural networks \citep{johnson2016composing}, while in traditional Bayesian models, the transformation tends to have a simple analytical form (e.g., the exponential function or inner product). One key challenge for Bayesian deep learning is on posterior inference, which is typically intractable for such models and needs sophisticated approximation techniques. Although much progress has been made on both variational inference and Monte Carlo methods~\citep{zhu2017big}, it is still too difficult for practitioners to pick-up. Moreover, although variational inference and Monte Carlo methods have their general recipes, it is still quite involved for an experienced researcher to derive the equations and implement every detail for a particular model. Such a procedure is error prone and takes a long time to debug. 

In this paper, we present ZhuSuan, a probabilistic programming library for Bayesian Deep Learning.\footnote{ZhuSuan is named after the Chinese name of abacus, which is the oldest calculating machine and has been recognized as the fifth greatest innovation in China.} We build ZhuSuan upon Tensorflow \citep{abadi2016tensorflow} to leverage its computation graphs for flexible modeling. With ZhuSuan, users can enjoy powerful fitting and multi-GPU training of deep learning, while at the same time they can use probabilistic models 
to model the complex world, exploit unlabeled data and deal with uncertainty by applying principled Bayesian inference. We provide running examples to illustrate how intuitive it is to program with ZhuSuan, including Bayesian logistic regression, variational auto-encoders, deep sigmoid belief networks and Bayesian recurrent networks. More examples can be found in our code repository \url{https://github.com/thu-ml/zhusuan}. The key difference of ZhuSuan from other probabilistic programming libraries lies in its flexibility, benefited from the deep learning paradigm on which the library is built, and the key treatment of model reuse. More detailed comparisons with two closely related works, Edward~\citep{tran2016edward} and PyMC3~\citep{salvatier2016probabilistic}, are provided in this paper.

\section{Bayesian Deep Learning}

We first overview the basics of Bayesian methods and their intersection with deep learning, and define the notations.

\subsection{Modeling} \label{sec:model}

As a conjunction of graph theory and probability theory, probabilistic graphical models (PGMs)~\citep{koller2009probabilistic} provide a powerful language for probabilistic machine learning. A PGM defines a joint distribution of a set of random variables using a graph, which has an intuitive and compact semantic meaning to read out the conditional independence structures. There are two major types of PGMs, namely, directed ones (aka. Bayesian networks) and undirected ones (aka. Markov random fields, MRFs). Though Bayesian networks and MRFs have different expressiveness, sometimes one can transform a Bayesian network to an MRF or vice versa~\citep{koller2009probabilistic}. In ZhuSuan, we concentrate on Bayesian networks, which provide an intuitive data generation process~\citep{gelman2014bayesian}. This choice is also reflected in the literature of Bayesian deep learning. For example, while modern deep generative models make their first remarkable step with undirected graphs~\citep{hinton2006fast}, the use of directed models \citep{kingma2013auto} has dominated the area due to cheap generation and fast inference. 


\begin{figure}[h]
  \centering
  \subcaptionbox{\label{fig:bn-graph}}
    {\includegraphics[height=4cm]{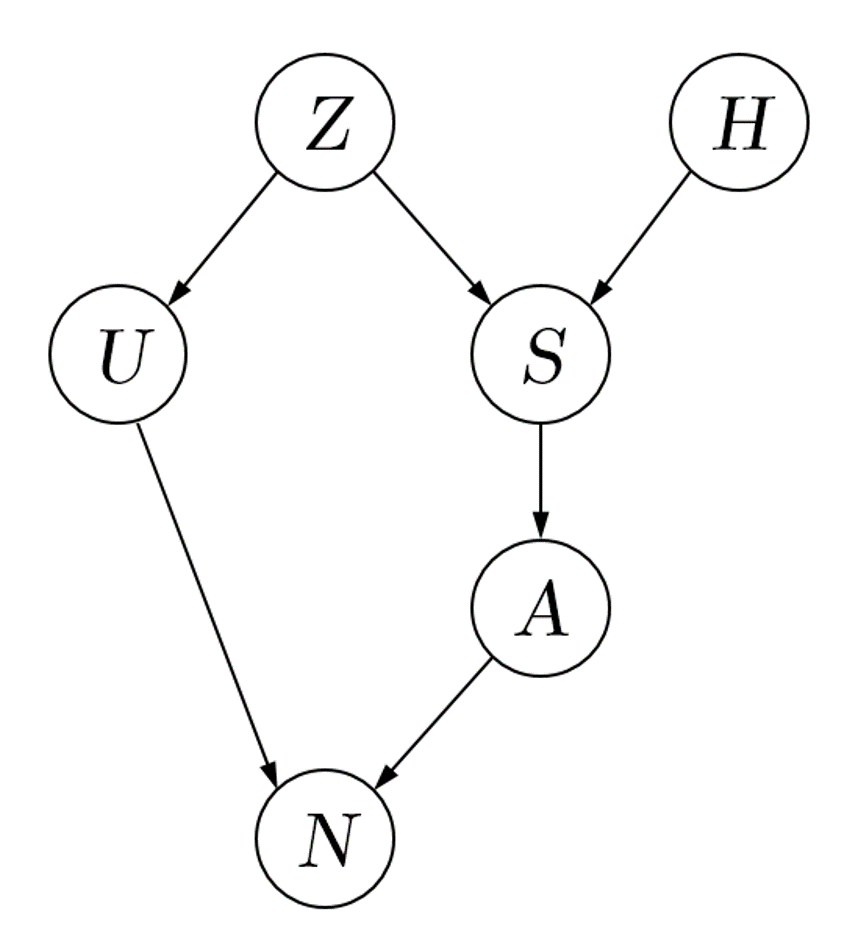}}
    \subcaptionbox{\label{fig:nodes}}
    {\includegraphics[height=4cm]{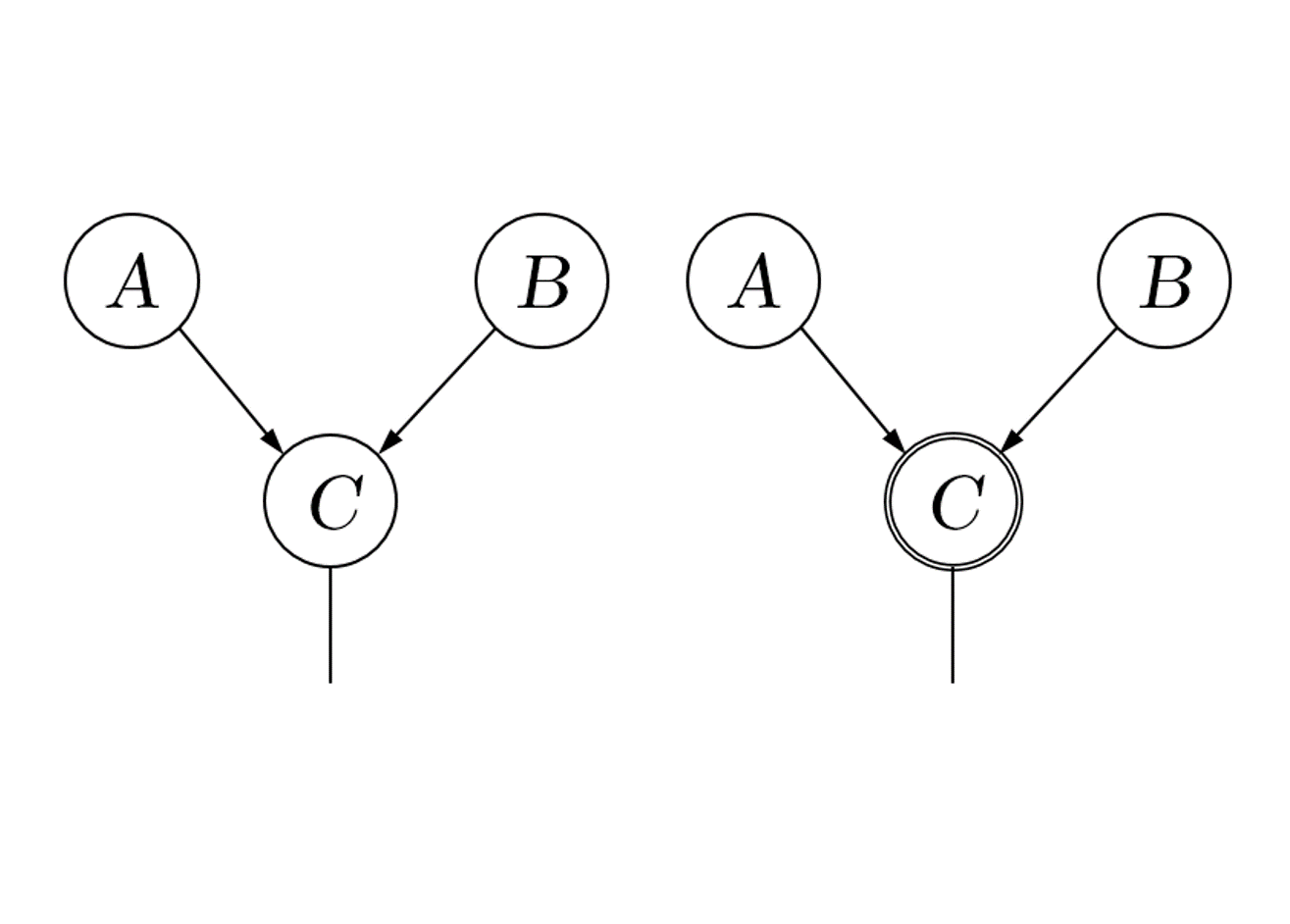}}
    \subcaptionbox{\label{fig:global-local}}
    {\includegraphics[height=4cm]{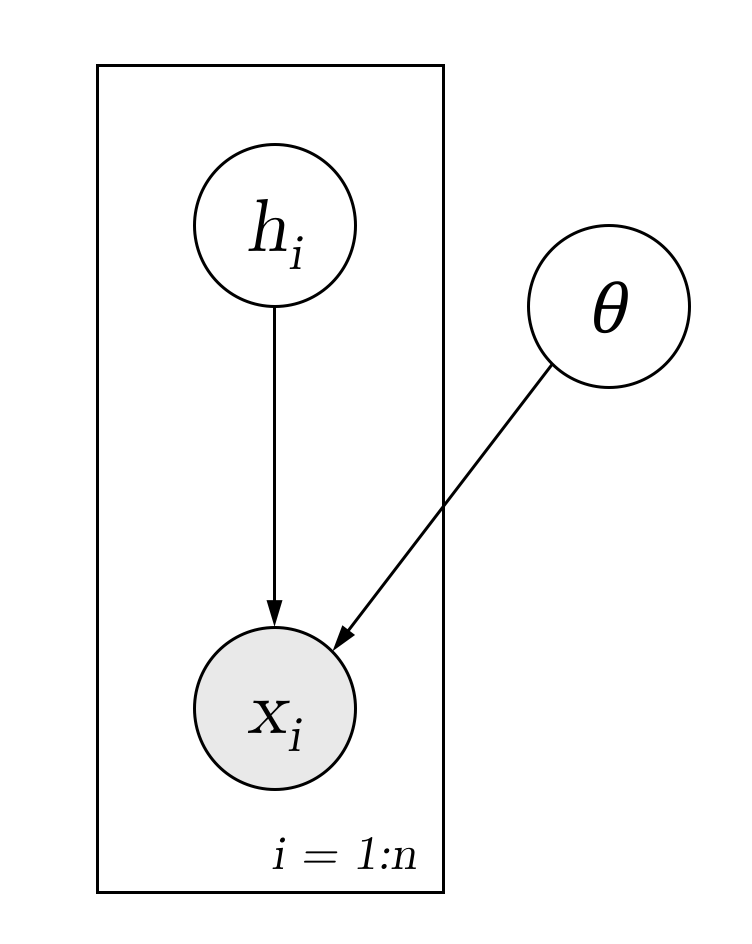}}
	\label{fig:bn}

    \caption{Bayesian networks}
\end{figure}

Figure~\ref{fig:bn-graph} shows a Bayesian network with a directed acyclic graph (DAG) structure. This represents a joint probability distribution over the nodes that factorizes according to the graph:
\begin{equation}
p(Z, H, U, S, A, N) = p(Z)p(H)p(U|Z)p(S|Z,H)p(A|S)p(N|U, A).
\end{equation}
The joint probability is the product of two types of factors: the prior distribution of variables at the roots (e.g., $p(Z)$) and the conditional probability distribution of other variables given their parents (e.g., $p(S|Z,H)$). The conditional distributions are also known as conditional probability tables (CPTs) for discrete variables and conditional probability density (CPD) functions for continuous variables. 

We consider the general formulation, where there are two kinds of nodes in a Bayesian network: {\it deterministic nodes} (Figure~\ref{fig:nodes}, right, double line notation) and {\it stochastic nodes} (Figure~\ref{fig:nodes}, left). A deterministic node calculates its value by a deterministic function of its parents,
while a stochastic node is represented by a standard probabilistic distribution. Although deterministic nodes are often not represented explicitly to make the graph structure concise, it is beneficial to do so especially when we consider Bayesian deep learning models. This is because the transformation in a traditional model is typically in a simple analytical form, while a Bayesian deep learning model learns it from a very flexible family (e.g., deep neural networks). Hence, explicitly representing deterministic nodes helps on both model definition and inference, as shall be clear soon. 

If all the variables are observed in the dataset, it is easy to do learning and inference for a Bayesian network. However, in reality it is more common to have partially observed data due to physical randomness, missing information and measurement noise, thereby some variables are hidden. Such models are known as latent variable models (LVMs), which provide a suite of useful tools to unveil the underlying factors (e.g., topics or latent feature representations). Figure~\ref{fig:global-local} illustrates a generic latent variable model, where the gray nodes represent the observed variables ($\bx_{1:n}$), the rest ($\bh_{1:n}, \theta$) are latent variables, and $n$ is the number of observed data samples. This class of models include both global and local latent variables. By \textit{local} we mean the latent variable only has impact on its paired observation (e.g., $\bh$ is paired with $\bx$), while \textit{global} means it influences all observations (e.g., $\theta$).

\subsection{Inference} \label{sec:infer}
For a latent variable model, posterior inference is the key step to unveil the latent factors for a particular input. To make the notations simple, for this subsection we consider a highly abstract model $p(\bz, \bx) = p(\bz)p(\bx|\bz)$, where $\bz$ represents all latent variables, including the global and local ones, i.e., $\bz = (\bh, \theta)$, and $\bx$ denotes observations. The vanilla Bayes' rule provides a principle to derive the posterior distribution:
\begin{eqnarray}
p(\bz | \bx) = \frac{p(\bz,\bx)}{p(\bx)} = \frac{p(\bz)p(\bx|\bz)}{p(\bx)}.
\end{eqnarray}
As analyzed in~\citep{zhu2014bayesian}, the vanilla Bayes' rule can be generalized to regularized Bayesian inference (RegBayes) by introducing posterior regularization under an information theoretical formulation. 

In general, the posterior inference using Bayes' rule or the RegBayes formulation is intractable, except a few simple examples. Therefore, we have to resort to approximate Bayesian inference methods. Many years of developments in this area has led to many fast, widely applicable approximate inference algorithms, mainly divided into two categories, variational inference and Monte Carlo methods~\citep{zhu2017big}. Both will be covered below.

\subsubsection{Variational Inference}\label{sec:vi}

Variational inference (VI) is an optimization-based method for posterior approximation, in which a parametric distribution family $q_{\phi}(\bz)$ is chosen to approximate the true posterior $p(\bz|\bx)$ by minimizing the KL divergence between them ($\mathrm{KL}(q_{\phi}(\bz)\|p(\bz|\bx)$).\footnote{We do not explicitly condition $\bz$ on $\bx$ in $q$ because it is not necessary to explicitly model this, though many modern literatures do. In fact, even without explicit modeling, the optimization process will connect them together.}
The KL-divergence minimization is equal to maximizing a lower bound of the marginal log likelihood $\log p(\bx)$:
\begin{equation} \label{eq:elbo}
\begin{aligned}
\mathcal{L}(\bx; \phi) &= \log p(\bx) - \mathrm{KL}(q_{\phi}(\bz)\|p(\bz|\bx)) \\
&= \mathbb{E}_{q_{\phi}(\bz)}\left[\log p(\bx|\bz)\right] - \mathrm{KL}\left(q_{\phi}(\bz)\|p(\bz)\right).
\end{aligned}
\end{equation}
In the VI literature, $q_{\phi}(\bz)$ is called the \textit{variational posterior}, and $\mathcal{L}(\bx; \phi)$ is called the \textit{evidence lower bound} (ELBO). Generally there are two steps for doing VI. First is to choose the parametric family that will be used as the variational posterior. Then the second step is to solve the optimization problem with respect to the variational parameters ($\phi$).

In recent years, benefited from the joint effort of the Bayesian and the deep learning community, variational inference is undergoing many significant changes, both in the algorithm and the choices of variational families. From the algorithm side, stochastic approximation with data sub-sampling has enabled VI to scale to large datasets \citep{hoffman2013stochastic}. Meanwhile, direct optimization of the variational lower bounds by gradient descent is replacing traditional analytic updates, which makes VI applicable to a broader range of models with non-conjugate dependencies \citep{graves2011practical,titsias2014doubly}. The key challenge that draws most attention in this direction is on the gradient estimates. Many gradient estimators have been developed with different variance reduction techniques \citep{paisley2012variational,MnihGregor2014,mnih2016variational}. Specially, for continuous latent variable models, an algorithm named \textit{Stochastic Gradient Variational Bayes} (SGVB) has been very successful \citep{kingma2013auto}, due to a clever trick to pass the gradient through a stochastic node, which is now well-known as \textit{the reparameterization trick}. Besides, better lower bounds than the ELBO have also been developed, which can be optimized by gradient descent as well and have been applied to both discrete and continuous latent variable models \citep{burda2015importance,mnih2016variational}. 

On the other side, considerable efforts have also been put into the design of variational posteriors. Using neural networks to parameterize the variational posterior has been a common practice adopted by aforementioned works \citep{MnihGregor2014,kingma2013auto}, serving as one of the key technologies to efficiently train deep generative models. The network is usually fed with observations and is expected to amortize the inference cost for each data point by learning to do inference. This type of scheme is often referred as \textit{amortized inference} \citep{rezende2015variational}. In summary, VI in the Bayesian deep learning context is more stochastic, differentiable, and amortized than before.

\subsubsection{Monte Carlo Methods}
\label{sec:mc}

Unlike VI that reformulates the inference as an optimization problem, Monte Carlo methods \citep{Robert:2005:MCS:1051451} are more direct ways to simulate samples or estimate properties of a particular distribution. Two main techniques extensively used for Bayesian inference are Importance Sampling and Markov Chain Monte Carlo.

\textbf{Importance Sampling} The basic idea of importance sampling is as follows. To estimate $\mu = \mathbb{E}_{p} f(\bx)$ where $p$ is a probability density defined over $\mathbb{R}^d$, a probability distribution $q$ called the proposal is introduced. It is required that $q(\bx) > 0$ whenever $p(\bx)f(\bx) > 0$. Then
\begin{equation} \label{eq:is}
\mu = \int_{\mathbb{R}^d} f(\bx)p(\bx)\;d\bx = \int_{\mathbb{R}^d} \frac{f(\bx)p(\bx)}{q(\bx)}q(\bx)\;d\bx = \mathbb{E}_{q}\left[\frac{f(\bx)p(\bx)}{q(\bx)}\right].
\end{equation}
The Monte Carlo estimate of $\mu$ given by importance sampling is
\begin{equation} \label{eq:is-est}
\hat{\mu} = \frac{1}{N}\sum_{i=1}^N \frac{f(\bx_i)p(\bx_i)}{q(\bx_i)},\quad\bx_i \sim q(\bx).
\end{equation}
In the Bayesian inference context, it is often the case that we only have access to an unnormalized version of $p$, which we denote as $\tilde{p}$. $p = \tilde{p}/Z$, where $Z$ is the normalizing constant which is intractable. \textit{Self-normalized Importance Sampling} is particularly useful in this situation, which gives the estimate
\begin{equation} \label{eq:nis-est}
\tilde{\mu} = \sum_{i=1}^N \tilde{w}_i f(\bx_i),\quad\text{where} \quad\tilde{w}_i = \frac{w_i}{\sum_{j=1}^N w_j}, w_i = \frac{\tilde{p}(\bx_i)}{q(\bx_i)}.
\end{equation}
It can be proved that $\tilde{\mu}$ is an asymptotically unbiased estimator of $\mu$ as $N\to\infty$. For more detailed introduction, we refer the readers to \citep{mcbook}.

As introduced above, importance sampling in the strict sense is not a sampling method because it does not directly draw samples from the target distribution. Instead, it provides a method for estimating properties of a certain distribution (normalized or not). Thus the use of importance sampling can be everywhere that needs an estimate of some integral over a probability measure. Though the computation in \cref{eq:is-est,eq:nis-est} is rather direct, some of their application scenarios may not be obvious for users.

Here we focus on application scenarios in Bayesian deep learning, which concentrate on model learning and evaluation. For model learning, it is shown in \citep{bornschein2014reweighted} that self-normalized importance sampling can be used to estimate gradients of marginal log likelihoods with respect to model parameters, where this technique was proposed to train deep generative models. Even if not used for learning, importance sampling is still a simple and efficient method for evaluating marginal log likelihoods of latent variable models \citep{wallach2009evaluation}. One of the drawbacks of using importance sampling is that the estimation has large variance if the proposal is not good enough \citep{mcbook}. Therefore people have been exploring the use of adaptive proposals \citep{cappe2008adaptive, cornuet2012adaptive}. This idea is recently reformed into a new technique called neural adaptive proposals, i.e., a neural network is used to parameterize the proposal distribution, which is adapted towards the optimal proposal by gradient descent \citep{bornschein2014reweighted}. This technique proves to be successful in applications like dynamic systems \citep{gu2015neural}, inference acceleration \citep{7961239,paige2016inference} and learning attention models \citep{ba2015learning}.

\textbf{Markov Chain Monte Carlo} 
Markov Chain Monte Carlo (MCMC) is a classic method to generate samples from the posterior distribution in Bayesian statistics. Unlike variational inference, MCMC is known to be asymptotically unbiased and allows the user to trade off computation for accuracy without limit. 
The basic idea of MCMC is to design a Markov chain whose stationary distribution is the target distribution, then samples can be generated by simulating the chain until convergence.  
Specifically, for a Markov chain specified by the transition kernel $T(\bz'|\bz)$, the most common sufficient condition for $p(\bz|\bx)$ being its stationary distribution is the detailed balance condition:
$$
p(\bz|\bx)T(\bz^\prime|\bz, \bx)=p(\bz^\prime|\bx)T(\bz|\bz^\prime, \bx),\quad \mbox{for all }\bz, \bz^\prime.
$$
An arbitrarily chosen initial state $\bz^0$ and the transition kernel define a joint distribution $p(\bz^0, \dots, \bz^t, \dots|\bx)$ and marginal distributions $p(\bz^t|\bx)$. It can be shown that under mild conditions, $p(\bz^t|\bx)$ converges to $p(\bz|\bx)$ as $t\rightarrow \infty$~\citep{Robert:2005:MCS:1051451}. In practice, it is common to throw away samples from the initial stage before the chain converges. This stage is often referred as the \emph{burn-in} (or \emph{warm-up}) stage. Once we get samples from the MCMC methods, there are many ways to use them. For parameter estimation of LVMs, the samples can be used in the Monte-Carlo EM (MCEM) algorithm~\citep{wei1990monte}.

Throughout the literature there are many ways to design the transition kernel. For example, the simplest form of the Metropolis Hastings algorithm~\citep{metropolis1953equation,hastings1970monte} combines a Gaussian random walk proposal with an accept-reject test for correction, which scales poorly with increasing dimension and complexity of the target distribution. Gibbs sampling~\citep{geman1984stochastic} utilizes the structure of the target distribution by taking its element-wise conditional distribution as the transition proposal. However, it requires the conditionals to be analytically computable, which limits its application scope, especially in Bayesian deep learning.

A powerful MCMC method that efficiently explores high-dimensional continuous distributions is \emph{Hamiltonian Monte Carlo} (HMC)~\citep{neal2011mcmc}.
To sample from $p(\bz|\bx)$, HMC introduces an auxiliary momentum variable $\bp$ with the same dimensionality as $\bz$, and effectively samples from the joint distribution
$p(\bz, \bp|\bx)=p(\bz|\bx)\exp(-\frac{1}{2}\bp^\top M^{-1}\bp)$, where $M$ is called the mass matrix.
Samples are generated by simulating the \emph{Hamiltonian dynamics}, which governs the evolution of the $(\bz, \bp)$ system along continuous time $t$ as:
\begin{equation}
\frac{\partial \bz}{\partial t} = \nabla_{\bp} H,\quad \frac{\partial \bp}{\partial t} = -\nabla_{\bz} H.
\end{equation}
Here $H(\bz, \bp) = -\log p(\bz,\bp|\bx)$ is the \emph{Hamiltonian} of the system.
The Hamiltonian dynamics defines a continuous-time transition kernel, and the stationary distribution of the corresponding Markov chain is the desired $p(\bz, \bp|\bx)$, due to the volume-preservation and the Hamiltonian-conservation properties of the dynamics.
In practice one has to simulate the dynamics in discrete time steps. The leapfrog integrator~\citep{leimkuhler2004simulating} is widely used, since it keeps the two vital properties of the Hamiltonian dynamics: it is volume-preserving and approximately Hamiltonian-conserving. Combined with a Metropolis-Hastings algorithm to correct the approximating error of the Hamiltonian, the discrete-time  simulation comes to our satisfactory: the Markov chain converges to our desired distribution.
The key advantage of HMC is that it exploits information about the geometry of the target distribution \citep{betancourt2017conceptual} while only needs the distribution density and the gradient to proceed.

People have been exploring the use of HMC in Bayesian deep learning. The early work by Radford Neal~\citep{neal1995bayesian} applied HMC to the inference of Bayesian neural networks, one of the representative models that apply Bayesian methods to capture uncertainty in deep learning. Recent works on deep generative models have applied HMC to improve the variational posterior~\citep{salimans2015markov} as well as used HMC-based MCEM algorithm to directly learn the model parameters~\citep{pmlr-v70-hoffman17a,li2017approximate,titsias2017learning}.

\section{Design Overview}

ZhuSuan has been designed towards the basic needs of Bayesian deep learning, which, as stated in the last section, mainly include two parts: modeling and inference. As for the modeling part, we follow the principle that the code reads like the model definition. This requires model primitives that support:
\begin{itemize}
\item Direct building of probabilistic models as computation graphs.
\item Easy reuse by changing the states of stochastic nodes (observed or not).
\item Arbitrary deterministic modeling with the full power of Tensorflow.
\end{itemize}
On the side of inference, to leverage recent advances in Bayesian deep learning while staying applicable to a broad class of models, ZhuSuan makes efforts towards:
\begin{itemize}
\item Unifying traditional and advanced differentiable inference methods in a deep learning paradigm.
\item Supporting inference for both continuous latent variables and discrete ones.
\end{itemize}
Also from users' perspective, the design of ZhuSuan has been governed by two principles \footnote{Both principles are learned from Lasagne \citep{lasagne}.}:
\begin{itemize}
\item Modularity: Make flexible abstractions that allow all parts to be used independently.
\item Transparency: Do not hide the whole inference procedure behind abstractions, to enable heavy customization by users.
\end{itemize}
This makes ZhuSuan particularly different from existing probabilistic programming libraries, which will be discussed in later comparison (Section~\ref{sec:comp}).

\section{Features}

In this section we introduce the basic features of ZhuSuan, including model primitives and inference algorithms.

\subsection{Model Primitives}

For the reasons we have explained in Section~\ref{sec:model}, ZhuSuan's model primitives are designed for Bayesian networks. Since the basic structure is a DAG, we will introduce the node primitives first, and then will see how ZhuSuan stores and fetches information in the graph.

\textbf{Deterministic nodes} 
In ZhuSuan, users are enabled to use any Tensorflow operation for deterministic transformations. This includes various arithmetic operators (e.g., \longvar{tf.add, tf.tensordot, tf.matrix\_inverse}), neural network layers (e.g., \longvar{tf.layers.fully\_connected, tf.layers.conv2d}) and even control flows (e.g., \longvar{tf.while\_loop, tf.cond}). In Tensorflow computation graph, the output of an operation is named \longvar{Tensor}. ZhuSuan does not add any higher-level abstraction on \longvar{Tensor}s but directly treat them as deterministic nodes in the Bayesian networks. We will see that the other primitives work well directly with \longvar{Tensor}s.

\textbf{Stochastic nodes} For stochastic nodes, ZhuSuan provides an abstraction called \texttt{Stocha\allowbreak sticTensor}, which is named following its deterministic counterpart. Many commonly used probabilistic distributions are implemented and wrapped in \longvar{StochasticTensor}s (e.g., Normal, Bernoulli, Categorical), together with some recently developed  variants for Bayesian deep learning, such as Gumbel-softmax or Concrete \citep{jang2016categorical,maddison2016concrete}. \longvar{StochasticTensor}s inherit most behaviors of \longvar{Tensor}s. The former can be directly fed into Tensorflow operations and can automatically be cast into \longvar{Tensor}s when computing with them. When cast into 
\longvar{Tensor}s, the values they choose can be set to samples or observations, or can be automatically determined according to their states given in the context, which will be introduced next.

\textbf{The graph context} Above we see a Bayesian network can be built transparently with mixes of Tensorflow operations and \longvar{StochasticTensors}. Because large and sophisticated models are generally common today, it is still painful for one to deal with all these nodes individually. To help users manage the graph in a convenient way, ZhuSuan provides a graph context called the \longvar{BayesianNet}, which keeps track of all named \longvar{StochasticTensor}s constructed in it.

To start a \longvar{BayesianNet} context, use the \longvar{with} statement in python:
\begin{minted}{python}
import zhusuan as zs
with zs.BayesianNet() as model:
    # Build the graph.
\end{minted}
The \longvar{BayesianNet} context supports making queries on the inner stochastic nodes. The query options include the current-state outputs and local probabilities (the current-state value of CPDs) at certain nodes.

We use several representative examples to illustrate the process of building models in ZhuSuan, ranging from the simple Bayesian logistic regression to modern Bayesian deep models. In general, the programming of a probabilistic model on ZhuSuan is as intuitive as reading the corresponding graphical model, which is provided side-by-side.

\begin{examp}[Bayesian Logistic Regression, BLR] \label{exp:blr}
The generative process of a Bayesian logistic regression can be written as:
\begin{equation} \label{eq:blr}
\begin{aligned}
\bw &\sim N(\bzero, \alpha^2\bI), \\
y_i &\sim \mathrm{Bernoulli}(\sigma(\bw^T\bx_i)),\quad i = 1,\dots,n,
\end{aligned}
\end{equation}
where $\bw, \bx_i \in \mathbb{R}^D, y_i \in \{0, 1\}$, and $\sigma(\cdot)$ is the sigmoid function. The graphical model and the corresponding code is shown in Figure~\ref{fig:blr}. Note that data is modeled in a batch.
\begin{figure}[h]
  \begin{subfigure}[b]{.3\linewidth}
    \centering
    \includegraphics[width=0.8\textwidth]{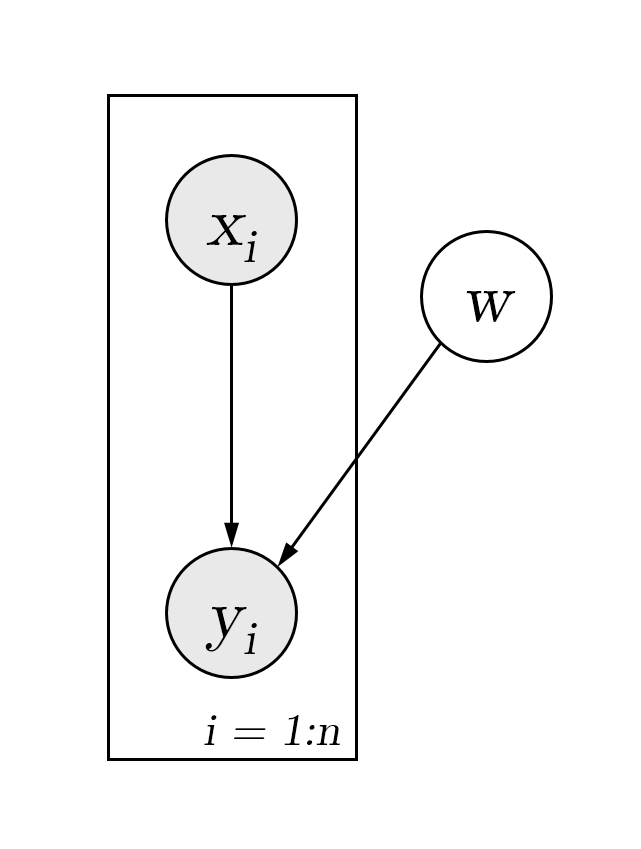}
    \caption{}
  \end{subfigure}
  \begin{subfigure}[b]{.7\linewidth}
    \centering
\begin{minted}{python}
import tensorflow as tf
with zs.BayesianNet() as model:
    # Define inputs and parameters
    x = tf.placeholder([None, D], tf.float32)
    log_alpha = tf.Variable(tf.zeros([D]))    
    # w ~ N(0, alpha^2*I)
    w = zs.Normal('w', mean=0., logstd=log_alpha, 
                  group_ndims=1)
    # y_logit = w^Tx
    y_logit = tf.reduce_sum(
        tf.expand_dims(w, 0)*x, axis=1)
    # y ~ Bernoulli(sigmoid(y_mean))
    y = zs.Bernoulli('y', y_logit)
\end{minted}
    \caption{}
  \end{subfigure}
  \caption{BLR: (a) graphical model; (b) programming on ZhuSuan.}
  \label{fig:blr}
\end{figure}

In this example the deterministic transformation part is the linear model ($\bw^T\bx$), which is implemented by the Tensorflow operations, \longvar{tf.expand\_dims, tf.reduce\_sum} and \longvar{tf.multiply} (\longvar{*}) to enable batch processing of the inputs ($\bx_i, i=1,\dots,n$). The two random variables $y$ and $\bw$ are created as \longvar{zs.Bernoulli} and \longvar{zs.Normal} respectively, which are the \longvar{StochasticTensor} for Bernoulli and normal distributions. The \longvar{group_ndims} argument passed to specify the \longvar{StochasticTensor} for $\bw$ means that the last dimension of $\bw$ is treated as a group, whose probabilities are computed together.
\end{examp}

\begin{examp}[Variational Auto-Encoders, VAE] \label{exp:vae}
Variational autoencoders (VAE) is one of the most popular products of Bayesian deep learning. The generative process of a VAE for modeling binarized MNIST data is
\begin{equation} \label{eq:vae}
\begin{aligned}
\bz &\sim N(\bzero, \bI), \\
\bx_{logits} &= f_{NN}(\bz), \\
\bx &\sim \mathrm{Bernoulli}(\sigma(\bx_{logits})),
\end{aligned}
\end{equation}
where $\bz \in \mathbb{R}^D, \bx \in \mathbb{R}^{784}$. This generative process is a stereotype for deep generative models, which starts with a latent representation ($\bz$) sampled from a simple distribution (such as the standard normal). Then the samples are forwarded through a deep neural network ($f_{NN}$) to capture the complex generative process of high dimensional observations such as images ($\bx$). Finally, some noise is added to the output to get a tractable likelihood for the model. For binarized MNIST, the observation noise is chosen to be Bernoulli, with its parameters output by the neural network. Because ZhuSuan is built upon Tensorflow, which has full support for deep neural networks, the implementation of VAE is extremely easy and intuitive, as shown in Figure~\ref{fig:vae}.
\begin{figure}[h]
  \begin{subfigure}[b]{.25\linewidth}
    \centering
    \includegraphics[width=0.8\textwidth]{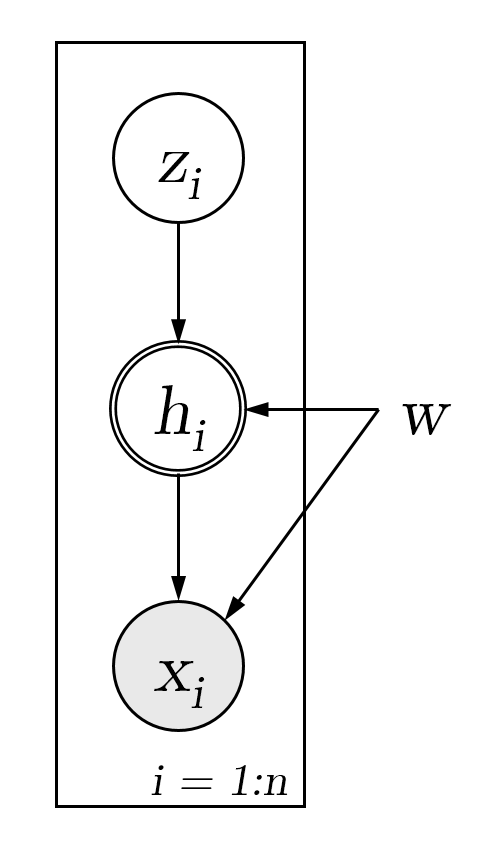}
    \caption{}
  \end{subfigure}
  \begin{subfigure}[b]{.75\linewidth}
    \centering
\begin{minted}{python}
from tensorflow.contrib import layers
with zs.BayesianNet() as model:
    z_mean = tf.zeros([n, D])
    z = zs.Normal('z', z_mean, std=1., group_ndims=1)
    h = layers.fully_connected(z, 500)
    x_logits = layers.fully_connected(h, 784, 
                                      activation_fn=None)
    x = zs.Bernoulli('x', x_logits, group_ndims=1)
\end{minted}
    \caption{}
  \end{subfigure}
  \caption{VAE: (a) graphical model; (b) programming on ZhuSuan.}
  \label{fig:vae}
\end{figure}
\end{examp}

\begin{examp}[Deep Sigmoid Belief Networks, DSBN] \label{exp:dsbn}
Sigmoid Belief Networks (SBN) is a directed discrete latent variable model that has close connections with feed-forward neural networks and Boltzmann Machines \citep{neal1992connectionist}. In recent years, the return of neural networks has brought a new life to this old model. In fact, the well-known Deep Belief Networks (DBN), the earliest work on deep learning, is an infinite-layer tied-weight SBN with the bottom layers untied \citep{hinton2006fast}. The generative process of a DSBN with $L$ layers is
\begin{equation}
\begin{aligned}
\bz^{(L)} &\sim \mathrm{Bernoulli}(\sigma(\bp^{(L)})), \\
\bz^{(l - 1)} &\sim \mathrm{Bernoulli}(\sigma(\bw^{(l)T}\bz^{(l)})),\quad l=L,\dots,1 \\
\bx &= \bz^{(0)},
\end{aligned}
\end{equation}
where $\sigma$ is the sigmoid function; $\bp^{(L)}$ is
the top layer parameters; $\bw$ are hidden weights; $\bz$ are latent variables, and $\bx$ are observations. From the definition we can see that $DSBN$ is a model with multi-layer stochastic nodes. The implementation of a two-layer DSBN ($L=2$) is shown in Figure~\ref{fig:sbn}.
\begin{figure}[h]
  \begin{subfigure}[b]{.25\linewidth}
    \centering
    \includegraphics[width=0.8\textwidth]{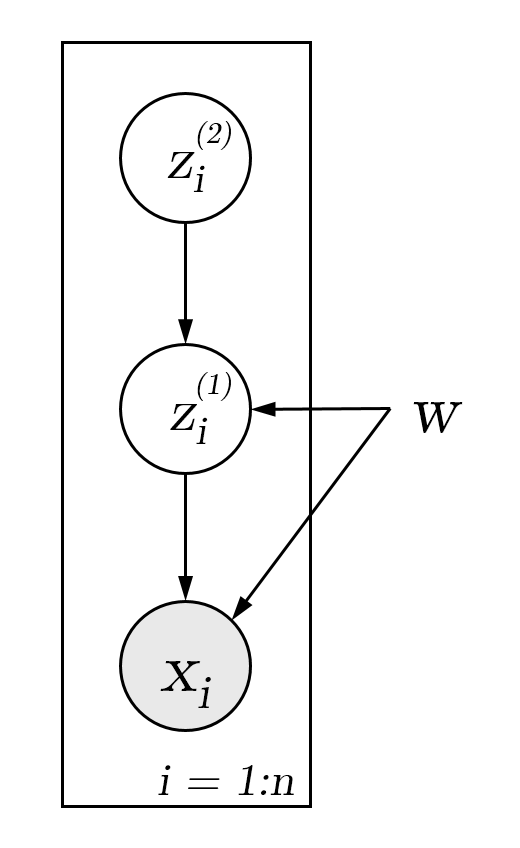}
    \caption{}
  \end{subfigure}
  \begin{subfigure}[b]{.75\linewidth}
    \centering
\begin{minted}{python}
with zs.BayesianNet() as model:
    z2_logits = tf.zeros([n, n_z])
    z2 = zs.Bernoulli('z2', z2_logits, dtype=tf.float32, 
                      group_ndims=1)
    z1_logits = layers.fully_connected(z2, n_z, 
                                       activation_fn=None)
    z1 = zs.Bernoulli('z1', z1_logits, dtype=tf.float32,
                      group_ndims=1)
    x_logits = layers.fully_connected(z1, n_x, 
                                      activation_fn=None)
    x = zs.Bernoulli('x', x_logits, group_ndims=1)
\end{minted}
    \caption{}
  \end{subfigure}
  \caption{DSBN: (a) graphical model; (b) programming on ZhuSuan.}
  \label{fig:sbn}
\end{figure}
\end{examp}

To provide a more interesting example that utilizes the powerful control flow operations from Tensorflow, we describe a \emph{Bayesian Recurrent Neural Network} (Bayesian RNN) below, which can also be intuitively programmed on ZhuSuan.

\begin{examp}[Bayesian RNN] \label{exp:brnn}
We have mentioned previously that deterministic neural networks lack the ability to account for the uncertainty of its own predictions. A solution to this given by Bayesian methods is a model named \emph{Bayesian Neural Network} (Bayesian NN), which treats the network weights as random variables and infers a posterior distribution over them given data. The generative process of a Bayesian NN for classification tasks is
\begin{equation}
\begin{aligned}
\bW &\sim N(\bzero, \bI), \\
\pi &= f_{NN}(\bx; \bW), \\
y &\sim \mathrm{Cat}(\mathrm{softmax}(\pi)),
\end{aligned}
\end{equation}
where $f_{NN}$ is a neural network with $\bW$ as the weights, $\pi$ is the predicted unnormalized log probabilities for all classes, $\mathrm{Cat}$ represents the categorical distribution, and $y$ is the class label. When $f_{NN}$ is chosen to be a recurrent network, the above process describes a Bayesian RNN. The graphical model for a Bayesian RNN is shown in Figure~\ref{fig:brnn}. Below we consider a model for a two-class sequence classification task. For the RNN part, it uses a Long Short-Term Memory (LSTM) network.
\begin{figure}[h]
    \centering
    \includegraphics[width=0.6\textwidth]{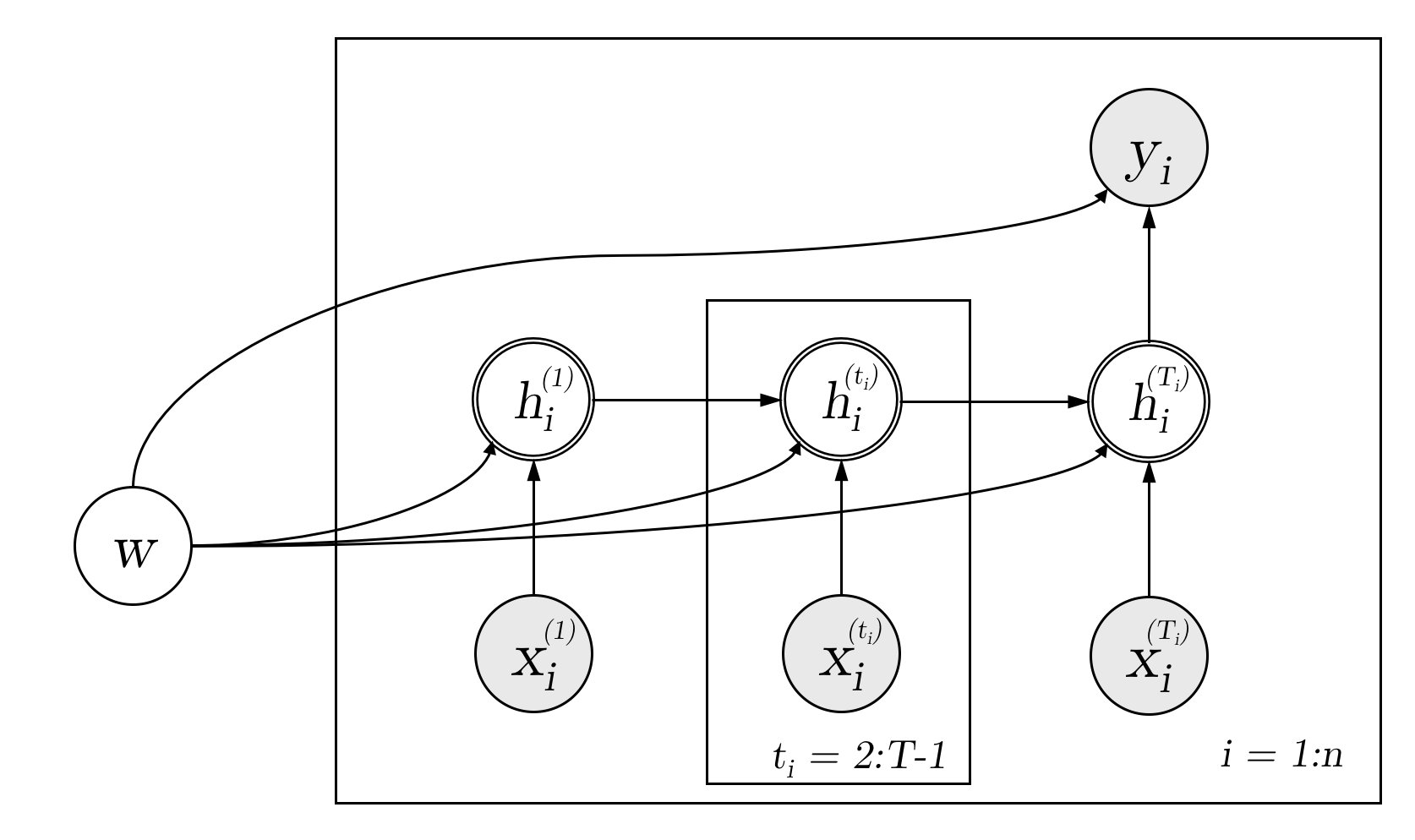}
    \captionof{figure}{Bayesian RNN}
    \label{fig:brnn}
\end{figure}%

\begin{minted}{python}
class BayesianLSTMCell(object):
    def __init__(self, num_units, forget_bias=1.0):
        self._forget_bias = forget_bias
        w_mean = tf.zeros([2 * num_units + 1, 4 * num_units])
        self._w = zs.Normal('w', w_mean, std=1., group_ndims=2)

    def __call__(self, state, inputs):
        c, h = state
        batch_size = tf.shape(inputs)[0]
        linear_in = tf.concat([inputs, h, tf.ones([batch_size, 1])], axis=1)
        linear_out = tf.matmul(linear_in, self._w)

        # i = input_gate, j = new_input, f = forget_gate, o = output_gate
        i, j, f, o = tf.split(value=linear_out, num_or_size_splits=4, axis=1)

        new_c = (c * tf.sigmoid(f + self._forget_bias) +
                 tf.sigmoid(i) * tf.tanh(j))
        new_h = tf.tanh(new_c) * tf.sigmoid(o)
        return new_c, new_h

def bayesian_rnn(cell, inputs, seq_len):
    initializer = (tf.zeros([batch_size, 128]), tf.zeros([batch_size, 128]))
    c_list, h_list = tf.scan(cell, inputs, initializer=initializer)
    relevant_outputs = tf.gather_nd(
        h_list, tf.stack([seq_len - 1, tf.range(batch_size)], axis=1))
    logits = tf.squeeze(tf.layers.dense(relevant_outputs, 1), -1)
    return logits

with zs.BayesianNet() as model:
    cell = BayesianLSTMCell(128, forget_bias=0.)
    logits = bayesian_rnn(cell, self.x, self.seq_len)
    _ = zs.Bernoulli('y', logits, dtype=tf.float32)
\end{minted}
The code splits into three parts. First we build a \longvar{BayesianLSTMCell} that applies transformations to the inputs and the hidden states at each time step. Its only difference from the \longvar{tf.nn.rnn_cell.BasicLSTMCell} in Tensorflow is that the weights are generated from a normal \longvar{StochasticTensor}. Then the \longvar{tf.scan} operation is used to apply the cell function to input sequences. This part utilizes the power of control flows to deal with the variable-length sequences (\longvar{tf.scan} is internally implemented by the \longvar{tf.while_loop} operator). Finally, a Bernoulli \longvar{StochasticTensor} generates the class label given the outputs from the RNN.
\end{examp}

\textbf{Model reuse} Unlike supervised neural networks, a key feature of probabilistic graphical models is polymorphism, i.e., a stochastic node can have two states: observed or latent. This is a major difficulty when designing programming primitives. For example, consider the above VAE case. If \texttt{z} is a Tensor sampled from the Normal distribution, when the model is again used in a case where \texttt{z} is observed, the common practice in Tensorflow programming is to write another piece of code that builds a graph with the observation of \texttt{z} as the input. When the set of stochastic nodes is large, the process is very painful.

The reusability problem is a shared problem for probabilistic programming libraries that are based on computation graph toolkits, e.g., Theano \citep{theano-2016} and Tensorflow \citep{abadi2016tensorflow}. Other libraries such as PyMC3 \citep{salvatier2016probabilistic} and Edward \citep{tran2016edward} address this problem by directly manipulating the created computation graph. Specifically, PyMC3 (based on Theano) relies on the officially supported \texttt{theano.clone()} function to copy and recreate subgraphs that are influenced when the state of a node changes. While Edward (based on Tensorflow) has to implement their own \texttt{copy()} function by looking into nonpublic low-level APIs of the Tensorflow computation graph. As we will see later, the solutions by manipulating the created graphs induce problems and limitations for these libraries.

ZhuSuan has been carefully designed towards reusability, but does not rely on altering the created graphs. Specifically, an observation can be passed to \longvar{StochasticTensor} directly, which enables model reuse by repeated calls of a function with different arguments passed. Below is an illustration example:
\begin{minted}{python}
def build_model(x_obs=None):
    x = zs.Normal('x', observed=a_obs)
    return f(x)

# samples from the Normal distribution will be used for computation in f
out = build_model()
# x_obs will be used for computation in f
out = build_model(x_obs=x_obs)
\end{minted}

For models that have many stochastic nodes (e.g., \longvar{x1, ..., xN}), in principle the polymorphism can be achieved by
\begin{minted}{python}
def build_model(x1_obs=None, ..., xN_obs=None):
    ...
\end{minted}
However, in practice this is often found to be redundant and makes it hard for automatic inference algorithms to fit each model. This is where the \longvar{BayesianNet} context makes a big difference. \longvar{BayesianNet} accepts an argument named \longvar{observed}, which is a dictionary mapping from names of \longvar{StochasticTensor}s to their observations. Then the context will be responsible for determining the states of \longvar{StochasticTensor}s. This makes it easier for users to handle state changes of large sets of stochastic nodes, and more importantly, enables a unified form of model building functions that makes automatic inference possible. An example is shown below.

\begin{minted}{python}
def build_model(observed=None):
    with zs.BayesianNet(observed=observed) as model:
        z = zs.Normal('z')
        x = zs.Normal('x', f(z))
        ...
    return model

# No observation
m = build_model()
# Observe z and x
m = build_model({'z': z_obs, 'x': x_obs})
\end{minted}

\subsection{Inference Algorithms}

In Section~\ref{sec:infer} we have covered the major inference methods used in Bayesian deep learning. Although most of them are gradient-based, stochastic, and black-box \citep{ranganath2014black}, which is suitable for building a general inference framework, there is currently no software that provides a complete support of them. To bridging the gap, ZhuSuan leverages recent advances of differentiable inference in Bayesian deep learning, and provides a wide and flexible support for both traditional and modern inference algorithms. These algorithms are provided in an API that fits well into deep learning paradigms, which makes inference as easy as doing gradient descent in deterministic neural networks. Below, we use examples to explain both variational inference and Monte Carlo methods in ZhuSuan. 

\subsubsection{Variational Inference}
\label{sec:zhusuan-vi}

As reviewed in Section~\ref{sec:vi}, a variational inference (VI) algorithm consists of two parts: the construction of variational posteriors and the optimization of variational objectives. 

Typical VI implementations in probabilistic programming languages have mostly restricted themselves to simple variational posteriors. For example, the ADVI algorithm \citep{JMLR:v18:16-107} that serves as the main VI method in Stan \citep{JSSv076i01} uses only Gaussian variational posteriors. In contrast, ZhuSuan supports building very flexible variational posteriors by leveraging Bayesian networks. This opens up the door for rich variational families with user-specified dependency structures.

\begin{table}[t]
\centering
\begin{tabular}{m{2cm}m{2.2cm}m{6.1cm}m{3.1cm}}
\toprule
Objective & \makecell{Gradient\\estimator} & Supported latent-variable types & \makecell{Implementations\\in \longvar{zs.variational}} \\ 
\midrule
\multirow{2}{2cm}{ELBO}        & SGVB      & 
\vspace{-0.1em}
\begin{itemize}[leftmargin=*]
\setlength\itemsep{0.1em}
\item continuous and reparameterizable
\item Concrete relaxation of discrete
\end{itemize}
  & \longvar{elbo().sgvb}\\
\cmidrule{2-4}
& REINFORCE & all types & \longvar{elbo().reinforce} \\
\midrule
\multirow{2}{2cm}{Importance weighted objective}  & SGVB (IWAE)   & 
\vspace{-0.1em}
\begin{itemize}[leftmargin=*]
\setlength\itemsep{0.1em}
\item continuous and reparameterizable
\item Concrete relaxation of discrete
\end{itemize} & \longvar{iw_objective().sgvb}\\
\cmidrule{2-4}
& VIMCO & all types & \longvar{iw_objective().vimco} \\
\midrule
$\mathrm{KL(p\|q)}$       & RWS       & all types & \longvar{klpq().rws} \\
\bottomrule
\end{tabular}
\caption{Variational inference in ZhuSuan. Relevant references are SGVB~\citep{kingma2013auto}, REINFORCE~\citep{williams1992simple, MnihGregor2014}, IWAE~\citep{burda2015importance}, VIMCO~\citep{mnih2016variational}, and RWS~\citep{bornschein2014reweighted}.} \label{tab:vi}
\end{table}

As for the optimization side, as mentioned in Section~\ref{sec:vi}, many gradient-based variational methods have emerged in recent progress of Bayesian deep learning \citep{kingma2013auto,MnihGregor2014,burda2015importance,mnih2016variational}. These methods differ in the variational objectives and the gradient estimators they use. To make them more automatic and easier to handle, ZhuSuan has wrapped them all into single functions, which computes a surrogate cost for users to directly take derivatives on. This means that optimizing these surrogate costs is equally optimizing the corresponding variational objectives using their well-developed gradient estimators. The currently supported variational methods are summarized in Table \ref{tab:vi}. They are grouped by the objective they aim to optimize. Currently there are three kinds of objectives supported: the ELBO (equivalent to $\mathrm{KL}(q\|p)$), the importance weighted objective \citep{burda2015importance} and $\mathrm{KL}(p\|q)$, where $p, q$ denotes the true and the variational posterior.

It is easy to program with ZhuSuan to implement a variational inference algorithm, following these steps:
\begin{enumerate}
\item[1)] Build the variational posterior using ZhuSuan's modeling primitives.
\item[2)] Get samples and their log probabilities from the variational posterior.
\item[3)] Provide the log joint function of the model and call variational objectives.
\item[4)] Choose the gradient estimator to use and get the surrogate cost to minimize.
\item[5)] Call Tensorflow optimizers to run gradient descent on the surrogate cost.
\end{enumerate}
Below we use a simple example to guide the readers through all these steps.

\begin{examp}[BLR, continued] We consider applying variational inference to the BLR model in Example~\ref{exp:blr}, following the above steps:

1) Build the variational posterior. The true posterior of $\bw$ is intractable and should have correlations across dimensions. Here we follow the common practice to 
make the mean-field assumption that the variational posterior $q(\bw)$ is factorized, i.e., $q(\bw) = \prod_{d=1}^D q(\bw_d)$, where $D$ is the dimension of the weights. The code for using factorized normal distribution as the variational posterior is:
\begin{minted}{python}
with zs.BayesianNet() as variational:
    w_mean = tf.Variable(tf.zeros([D]))
    w_logstd = tf.Variable(tf.zeros([D]))
    w = zs.Normal('w', w_mean, logstd=w_logstd, group_ndims=1)
\end{minted}

2) Get samples and their log probabilities from the variational posterior. This is by querying the stochastic node \texttt{w} about its outputs and local log probabilities. 
\begin{minted}{python}
qw_samples, log_qw = variational.query(
    'w', outputs=True, local_log_prob=True)
\end{minted}

3) Provide the log joint function of the model and call variational objectives. The log joint value can be computed easily by taking sum of log probabilities at individual nodes of the model:
\begin{minted}{python}
def log_joint(observed):
    model = blr(observed, x)
    log_py_xw, log_pw = model.local_log_prob(['y', 'w'])
    return log_py_xw + log_pw
\end{minted}
Then we call the ELBO objective, passing the log joint function, observed data and outputs of the variational posterior.
\begin{minted}{python}
lower_bound = zs.variational.elbo(log_joint, 
                                  observed={'y': y}, 
                                  latent={'w': [qw_samples, log_qw]})
\end{minted}

4) Choose the gradient estimator to use and get the surrogate cost to minimize. Because $\bw$ is continuous and can be reparameterized as $\bw = \epsilon\sigma_{\bw} + \mu_{\bw}$, where $\mu_{\bw}$ and $\sigma_{\bw}$ are the mean and the standard deviation of the normal distribution, we choose the SGVB gradient estimator (see Table~\ref{tab:vi} for the scope of application of each estimator). The computed surrogate costs are for a batch of data, which is then averaged.
\begin{minted}{python}
cost = tf.reduce_mean(elbo.sgvb())
\end{minted}

5) Call Tensorflow optimizers to run gradient descent on the surrogate cost. As previously explained, this will be optimizing the ELBO objective with the SGVB gradient estimator. We can also fetch the ELBO value to verify it.
\begin{minted}{python}
optimizer = tf.train.AdamOptimizer(learning_rate=0.001)
infer_op = optimizer.minimize(cost)
with tf.Session() as sess:
    for i in range(iters):
        _, elbo_value = sess.run([infer_op, elbo])
\end{minted}
\end{examp}

The above is a very simple example using factorized variational posteriors, while as mentioned previously, ZhuSuan supports building very flexible variational posteriors. In the following example, we will see how to do amortized inference (see Section~\ref{sec:vi}) by leveraging neural networks in the variational posterior.

\begin{examp}[VAE, continued]
Consider the VAE model in Example~\ref{exp:vae}. The key difference compared to BLR (Example~\ref{exp:blr}) is that VAE's latent variables $(\bz)$ are local instead of global. As the number of local latent variables scales linearly with the number of observations $(\bx)$, it is hard to afford a separate variational posterior for each of them. This is where the amortized style of inference becomes useful. Specifically, we use a neural network with $\bx$ as input to generate parameters of the variational posterior for each corresponding $\bz$. This network is often referred as \emph{decoder} or \emph{recognition model} in VAE. The code is shown in Figure~\ref{fig:vae-post}.
\begin{figure}[h]
  \begin{subfigure}[b]{.23\linewidth}
    \centering
    \includegraphics[width=0.8\textwidth]{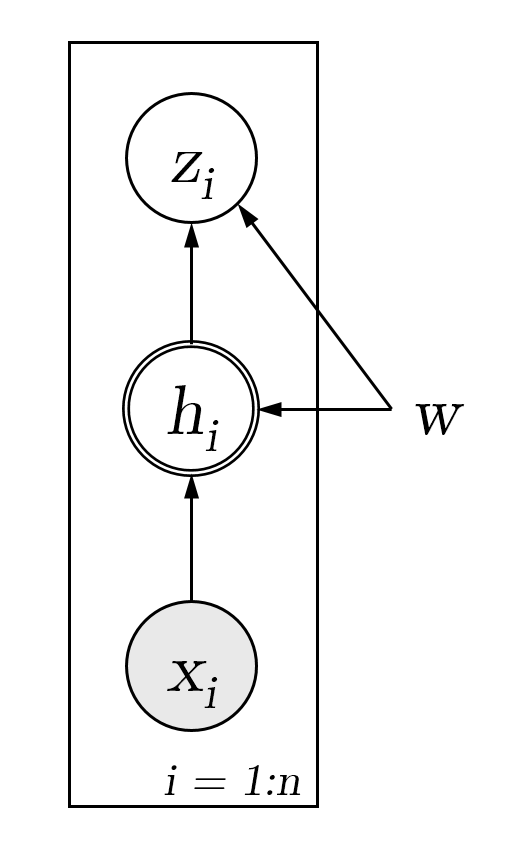}
    \caption{}
  \end{subfigure}
  \begin{subfigure}[b]{.77\linewidth}
    \centering
\begin{minted}{python}
with zs.BayesianNet() as variational:
    h = layers.fully_connected(x, n_h)
    z_mean = layers.fully_connected(h, n_z, 
                                    activation_fn=None)
    z_logstd = layers.fully_connected(h, n_z, 
                                      activation_fn=None)
    z = zs.Normal('z', z_mean, logstd=z_logstd,
                  group_ndims=1)
\end{minted}
    \caption{}
  \end{subfigure}
  \caption{VAE's variational posterior: (a) graphical illustration; (b) programming on ZhuSuan.}
  \label{fig:vae-post}
\end{figure}

Having built the variational posterior, we can follow the above steps to finish the inference procedure. Note that because the variational distribution for $\bz$ is again normal, we can use the SGVB gradient estimator for optimizing the ELBO objective. We omit these steps because they are very similar to those illustrated in the previous example.
\end{examp}

In deep, hierarchical models, there are often conditional dependencies between latent variables that need consideration during inference. We use the DSBN model from Example~\ref{exp:dsbn} to illustrate how to  
introduce structured dependencies when building variational posteriors for hierarchical models. This example also demonstrates applying VI to discrete latent variables in ZhuSuan. 

\begin{examp}[DSBN, continued] \label{exp:dsbn-vi}
A commonly used variational posterior in recent works for the DSBN model is
\begin{equation}
q(\bz^{(1:L)}) = \prod_{l=2}^L q(\bz^{(l)}|\bz^{(l-1)})q(\bz^{(1)}|\bx).
\end{equation}
which has the same conditional dependency structure with the original model given $\bx$ is observed. The graphical model as well as the code for this structured posterior (when $L=2$) are shown in Figure~\ref{fig:sbn-post}.
\begin{figure}[h]
  \begin{subfigure}[b]{.23\linewidth}
    \centering
    \includegraphics[width=0.8\textwidth]{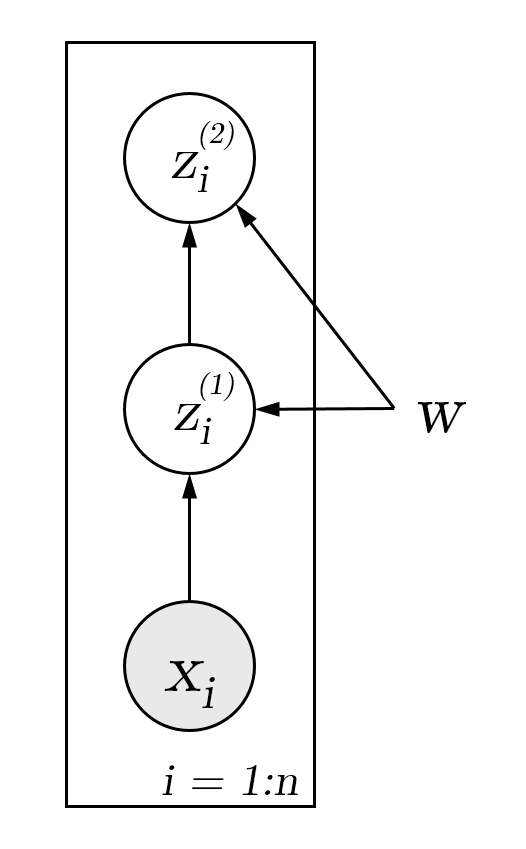}
    \caption{}
  \end{subfigure}
  \begin{subfigure}[b]{.77\linewidth}
    \centering
    \begin{minted}{python}
with zs.BayesianNet() as variational:
    z1_logits = layers.fully_connected(x, n_z, 
                                       activation_fn=None)
    z1 = zs.Bernoulli('z1', z1_logits, dtype=tf.float32)
    z2_logits = layers.fully_connected(z1, n_z, 
                                       activation_fn=None)
    z2 = zs.Bernoulli('z2', z2_logits, dtype=tf.float32)
\end{minted}
    \caption{}
  \end{subfigure}
  \caption{DSBN's variational posterior: (a) graphical illustration; (b) programming on ZhuSuan.}
  \label{fig:sbn-post}
\end{figure}
It can be seen from the code that $\bz^{(2)}$ directly depends on $\bz^{(1)}$ by a fully connected layer in neural networks. Note that here $\bz^{(2)}$ and $\bz^{(1)}$ are discrete latent variables. In ZhuSuan there are generally two ways for doing variational inference with them.

One way is to use the Concrete relaxation, or Gumbel-softmax trick~\citep{maddison2016concrete,jang2016categorical}, namely, we use Concrete random variables to replace the Bernoulli ones during training. Due to the Concrete distribution is continuous and reparameterizable, we can keep on using the SGVB estimator. In test time, we switch back to the Bernoulli random variables using the same input parameters. Both the model and the posterior in this way are modified a little:
\begin{minted}{python}
# model with Concrete relaxation
with zs.BayesianNet() as model:
    z2_logits = tf.zeros([n, n_z])
    if is_training:
        z2 = zs.BinConcrete('z2', z2_logits, group_ndims=1)
    else:
        z2 = zs.Bernoulli('z2', z2_logits, group_ndims=1, dtype=tf.float32)
    z1_logits = layers.fully_connected(z2, n_z, activation_fn=None)
    if is_training:
        z1 = zs.BinConcrete('z1', z1_logits, group_ndims=1)
    else:
        z1 = zs.Bernoulli('z1', z1_logits, group_ndims=1, dtype=tf.float32)
    x_logits = layers.fully_connected(z1, n_x, activation_fn=None)
    x = zs.Bernoulli('x', x_logits, group_ndims=1)

# posterior with Concrete relaxation
with zs.BayesianNet() as variational:
    z1_logits = layers.fully_connected(x, n_z, activation_fn=None)
    if is_training:
        z1 = zs.BinConcrete('z1', z1_logits, group_ndims=1)
    else:
        z1 = zs.Bernoulli('z1', z1_logits, group_ndims=1, dtype=tf.float32)
    z2_logits = layers.fully_connected(z1, n_z, activation_fn=None)
    if is_training:
        z2 = zs.BinConcrete('z2', z2_logits, group_ndims=1)
    else:
        z2 = zs.Bernoulli('z2', z2_logits, group_ndims=1, dtype=tf.float32)
\end{minted}
The other approach is to directly use a gradient estimator that applies to discrete latent variables. This includes REINFORCE and VIMCO. Here we present the code for using VIMCO. Because VIMCO is optimizing the importance weighted bound, which is a multi-sample bound, the model and the variational posterior need to be changed to multi-sample versions as shown below.
\begin{minted}{python}
# model (multi-sample version)
with zs.BayesianNet() as model:
    z2_logits = tf.zeros([n, n_z])
    z2 = zs.Bernoulli('z2', z2_logits, group_ndims=1, dtype=tf.float32,
                      n_samples=n_samples)
    z1_logits = layers.fully_connected(z2, n_z, activation_fn=None)
    z1 = zs.Bernoulli('z1', z1_logits, group_ndims=1, dtype=tf.float32)
    x_logits = layers.fully_connected(z1, n_x, activation_fn=None)
    x = zs.Bernoulli('x', x_logits, group_ndims=1)

# posterior (multi-sample version)
with zs.BayesianNet() as variational:
    z1_logits = layers.fully_connected(x, n_z, activation_fn=None)
    z1 = zs.Bernoulli('z1', z1_logits, group_ndims=1, dtype=tf.float32,
                      n_samples=n_samples)
    z2_logits = layers.fully_connected(z1, n_z, activation_fn=None)
    z2 = zs.Bernoulli('z2', z2_logits, dtype=tf.float32)
\end{minted}
Note the \longvar{n_samples} argument added in the code. This time we call the VIMCO gradient estimator to get the surrogate cost:
\begin{minted}{python}
lower_bound = zs.variational.iw_objective(
    log_joint, observed={'x': x}, latent=qz_outputs, axis=0)
cost = tf.reduce_mean(lower_bound.vimco())
\end{minted}
All samples along the axis 0 will be used to obtain variance reduced gradient estimates.

\end{examp}

\subsubsection{Monte Carlo Methods}
\label{sec:zhusuan-mc}

In addition to variational inference, efforts have also been made towards unifying Monte Carlo methods in the Bayesian deep learning context. We have covered the basics in Section~\ref{sec:mc}. Below we walk through ZhuSuan's support for both importance sampling and a black-box MCMC method: Hamiltonian Monte Carlo.

\textbf{Importance Sampling} As reviewed in Section~\ref{sec:mc}, the major application scenarios for importance sampling in Bayesian deep learning include model learning and evaluation. For model learning, we have introduced that self-normalized importance sampling can be used to estimate the gradients of marginal log likelihoods with respect to model parameters \citep{bornschein2014reweighted}. It turns out that the gradients estimated in this way are equivalent to exact gradients of the importance weighted variational objective (mentioned in Section~\ref{sec:zhusuan-vi}) with respect to model parameters. So the model learning procedure with importance sampling can be implemented by directly optimizing the importance weighted objective (\longvar{zs.variational.iw_objective()}, see Table~\ref{tab:vi}) with respect to the model parameters:
\begin{minted}{python}
lower_bound = zs.variational.iw_objective(
    log_joint, observed, latent, axis=axis)
optimizer = tf.train.AdamOptimizer(learning_rate)
learning_op = optimizer.optimize(lower_bound, var_list=model_params)
\end{minted}
The only difference is that the meaning of the variational posterior has changed to the proposal distribution.

Besides learning model parameters, we have mentioned that importance sampling is extensively used for model evaluation. Towards this need, ZhuSuan's evaluation module provides an \longvar{is_loglikelihood()} function for estimating the marginal log likelihoods of any given observations using simple importance sampling:
\begin{minted}{python}
ll = zs.is_loglikelihood(log_joint, observed, latent, axis=axis)
\end{minted}

Until now we haven't covered how to build a proposal distribution for importance sampling. In fact, this procedure is exactly the same as that of building a variational posterior. With ZhuSuan's modeling primitives, neural adaptive proposals mentioned in Section~\ref{sec:mc} can be easy to implement by leveraging neural networks in the proposal distribution. Adapting the proposal for the above two scenarios is also straightforward, where the true posterior distribution is often an ideal choice. So the adaptation turns out a variational inference problem, which can be solved by choosing an appropriate method in Table~\ref{tab:vi}. Specially, when the $\mathrm{KL}(p\|q)$ objective is chosen and the gradient estimation is by importance sampling, this recovers the method used in Reweighted Wake-Sleep (RWS) \citep{bornschein2014reweighted}, which is why this estimator is named \longvar{rws} in ZhuSuan.
We illustrate the process of training DSBN by importance sampling in the following example.

\begin{examp}[DSBN, continued] In this example we reproduce the algorithm proposed in the Reweighted Wake-Sleep paper~\citep{bornschein2014reweighted}, which learns DSBN by importance sampling and a neural adaptive proposal. The code snippet below follows from the multi-sample version of the model and the posterior (now used as the proposal) in Example~\ref{exp:dsbn-vi}, and we omit the code to draw samples and compute their log probabilities from the proposal distribution.
\begin{minted}{python}
# learning model parameters
lower_bound = tf.reduce_mean(
    zs.variational.importance_weighted_objective(
        log_joint, observed={'x': x_obs}, latent=latent, axis=0))
model_grads = optimizer.compute_gradients(-lower_bound, model_params)

# adapting the proposal
klpq_obj = zs.variational.klpq(
    log_joint, observed={'x': x_obs}, latent=latent, axis=0)
klpq_cost = tf.reduce_mean(klpq_obj.rws())
klpq_grads = optimizer.compute_gradients(klpq_cost, proposal_params)

infer_op = optimizer.apply_gradients(model_grads + klpq_grads)
\end{minted}
We can see that the code clearly has two parts. The first part is for learning model parameters, by optimizing the importance weighted objective with respect to model parameters. The second part is for adapting the proposal, by minimizing the inclusive KL divergence between the true posterior (the ideal proposal) and the current proposal. As the training proceeds, the adaptation part will help maintain a good proposal, reducing the variance of the marginal log likelihood estimate by importance sampling.
\end{examp}

\textbf{Hamiltonian Monte Carlo} In Section~\ref{sec:mc} we have briefly analyzed existing MCMC methods and have identified HMC as a powerful tool to address the posterior inference problem in high-dimensional spaces and non-conjugate models, which is a perfect fit for Bayesian deep learning. However, in practice this algorithm involves lots of technical details and can be hard to implement in a fast and efficient way. Besides, despite all tuning parameters in HMC have clear physical meanings, it is still hard for users to tune them by hand because the optimal choice always depends on unknown statistics of the underlying distribution. For example, the mass matrix describes the variance of the underlying distribution, which is hard to know before we can draw samples from it.

In recent years there has been a rise of practical algorithms and high-performance softwares that target at these problems. The \emph{No-U-Turn Sampler}, or \emph{NUTS}~\citep{hoffman2014no} proposes to automatically determine the number of leapfrog steps. It also comes along with a dual averaging scheme for automatically tuning the step size. The HMC implementation in Stan~\citep{JSSv076i01} also includes a procedure that estimates the mass matrix from the samples drawn in the warm-up stage.

ZhuSuan has learned from the existing libraries and provides a fast, automatic, and deep-learning style implementation of HMC. Specifically, ZhuSuan's HMC has the following features:
\begin{itemize}
\item Support running multiple chains in parallel on CPUs or GPUs.
\item Provide options for automatically tuning parameters, including the step size and the mass matrix. The NUTS algorithm for determining leapfrog steps is not included because it's a recursive algorithm and each separate chain can have different leapfrog steps, thus hard to have a parallel implementation in static computation graphs.
\item The algorithm is provided in an API very similar to Tensorflow optimizers, which is illustrated in Figure~\ref{fig:hmc}. We hope it is easy to start with for people who are familiar with deep learning libraries.
\end{itemize}
\begin{figure}[h]
  \begin{subfigure}[b]{.5\linewidth}
    \centering
\begin{minted}{python}
z = tf.Variable(0.)

hmc = zs.HMC(step_size=1e-3,
             n_leapfrogs=10)

sample_op, hmc_info = hmc.sample(
    log_joint, observed={'x': x}, 
    latent={'z': z})

with tf.Session() as sess:
    for i in range(iters):
        _ = sess.run(sample_op)
\end{minted}
    \caption{Using HMC in ZhuSuan}
  \end{subfigure}
  \begin{subfigure}[b]{.5\linewidth}
    \centering
\begin{minted}{python}
z = tf.Variable(0.)

optimizer = tf.train.AdamOptimizer(
    learning_rate=1e-3)

optimize_op = optimizer.minimize(
    cost(z))


with tf.Session() as sess:
    for i in range(iters):
        _ = sess.run(optimize_op)
\end{minted}
    \caption{Using Tensorflow optimizers}
  \end{subfigure}
  \caption{ZhuSuan's HMC vs. Tensorflow optimizers}
  \label{fig:hmc}
\end{figure}

\section{Comparison} \label{sec:comp}

\begin{table}[t]
\centering
\begin{tabular}{p{0.15\textwidth}p{0.26\textwidth}p{0.26\textwidth}p{0.21\textwidth}}
\toprule
Features & PyMC3 & Edward & ZhuSuan \\ 
\midrule
Based on        & Theano      & Tensorflow  & Tensorflow \\ \addlinespace
GPU support  & \cmark   & \cmark  & \cmark \\ \addlinespace
Deterministic modeling   & Any Theano operation  & Control flows (\longvar{tf.while_loop, tf.cond}) are not properly handled   & Any Tensorflow operation \\ \addlinespace
Customizable variational posterior &  Only for reparameterizable settings  & \cmark & \cmark \\ \addlinespace
HMC: Parallel chains  &  \cmark    & \xmark  & \cmark \\ \addlinespace
Modularity  & Modeling and inference are tightly coupled & Modeling and inference are tightly coupled & All parts can be used independently \\ \addlinespace
Transparency  & \xmark & \xmark & \cmark \\
\bottomrule
\end{tabular}
\caption{Comparison between ZhuSuan and other python probabilistic programming libraries.} \label{tab:comp}
\end{table}

We compare ZhuSuan with the representatives of python probabilistic programming libraries, namely PyMC3 \citep{salvatier2016probabilistic} and Edward \citep{tran2016edward}. A detailed feature comparison is shown in Table \ref{tab:comp}. As seen from the table, all the three libraries build upon modern computation graph libraries and can transparently run on GPUs. The other features can be mainly divided into three categories: modeling capabilities, inference support as well as architecture and design.

\textbf{Modeling capabilities} All the three libraries use the primitives from the computation graph toolkits they base on and are designed for directed graphs (Bayesian networks). PyMC3 solves the model reuse problem during inference by \longvar{theano.copy()} to copy the related subgraphs. Edward also does this but does not rely on the official Tensorflow API. ZhuSuan avoids altering created graphs and build reuse on purely function reuse and context management. As a result, PyMC3 and ZhuSuan can correctly deal with control flow primitives like \longvar{theano.scan()} and \longvar{tf.while_loop()}, while Edward faces challenges when applying variational inference to this kind of models because it will require graph copying when replacing latent variables with samples from the variational posterior, which is troublesome for control flow operations given there is no official support.

\textbf{Inference support} As for the range of inference methods, the three libraries have different emphases. PyMC3 started from MCMC methods, which is demonstrated by its name. It has put much efforts into sampling algorithms and have made them applicable to a broad class of models in Bayesian statistics. On the other hand, PyMC3's support for variational inference is limited and specific to several independent algorithms (\longvar{ADVI}, \longvar{SVGD}, etc.).  Edward has a general inference abstraction and implements a large number of algorithms as subclasses. However, as the form of abstraction induces constraints, many of these implemented algorithms have limited behaviors and make too strong assumptions on the constructed models (e.g., \longvar{GANInference}).  ZhuSuan emphasizes more on the scenarios from Bayesian deep learning thus puts more efforts into modern differentiable algorithms that can be unified into the deep learning paradigm. Both Edward and ZhuSuan support customizable variational posteriors for all the VI methods while PyMC3 has only made this specific to the reparameterizable settings.

\textbf{Architecture and design} The design philosophy of ZhuSuan has been introduced in Section 3, which emphasizes two principles---modularity and transparency. These two principles are very different from the traditional designs of probabilistic programming, for which PyMC3 can be a stereotype. The PyMC3 programs strictly follow the form with model definition, inference function call, and predictive checks. Little customization can be made except in the model definition and provided inference options. Edward also roughly follows this framework but has made it more flexible by general programmable variational posteriors. However, the inference procedure is also hidden due to its reliance on lots of internal manipulations of the created computation graphs, which can be hard for plain users to understand. Both the libraries build their modeling and inference functionalities in a tightly coupled way. This implies that if the model cannot be described using their modeling primitives, then there is little possibility using their inference features. ZhuSuan draws benefits from deep learning paradigms and makes inference just as doing gradient descent on a cost term. And all parts in the library can be used independently. This proves to be much more transparent and compositional, especially in models with a large set of stochastic activities, various observation behaviors and deep hierarchies.

\section{Conclusions}

We have described ZhuSuan, a python probabilistic programming library for Bayesian deep learning, built upon Tensorflow. ZhuSuan bridges the gap between Bayesian methods and deep learning by providing deep learning style primitives and algorithms for building probabilistic models and applying Bayesian inference. Many examples are provided to illustrate the intuitiveness of programming with ZhuSuan. We have open-sourced ZhuSuan on GitHub\footnote{GitHub address: \texttt{https://github.com/thu-ml/zhusuan}}, aiming to accelerate research and applications of Bayesian deep learning.


\acks{We thank Haowen Xu for helpful discussions on the API design, Chang Liu for advices on paper writing and Shizhen Xu, Dong Yan for initial attempts to try ZhuSuan on multi-card and distributed systems. We would like to acknowledge support for this project
from the National 973 Basic Research Program of China (No. 2013CB329403), National NSF of China (Nos. 61620106010, 61322308, 61332007), the National Youth Top-notch Talents Support Program, the NVIDIA NVAIL program, and Tsinghua Tiangong Institite for Intelligent Technology. }




\vskip 0.2in
\bibliography{sample}

\end{document}